\documentclass[conference]{IEEEtran}
\IEEEoverridecommandlockouts
\usepackage{cite}
\usepackage{amsmath,amssymb,amsfonts}
\usepackage{algorithmic}
\usepackage{graphicx}
\usepackage{textcomp}
\usepackage{xcolor}
\usepackage{xspace}
\usepackage{url}
\def\BibTeX{{\rm B\kern-.05em{\sc i\kern-.025em b}\kern-.08em
    T\kern-.1667em\lower.7ex\hbox{E}\kern-.125emX}}

\newcount\Comments  
\usepackage{color}
\definecolor{darkgreen}{rgb}{0,0.5,0}
\definecolor{purple}{rgb}{1,0,1}
\definecolor{teal}{rgb}{0,0.4627,0.5804}
\newcommand{\kibitz}[2]{\ifnum\Comments=1\textcolor{#1}{#2}\fi}

\newcommand{\veloffset}{\ensuremath{v_{\mathit{offset}}}\xspace}
    
\begin{document}

\title{A Middle Way to Traffic Enlightenment
\thanks{This work is supported by the National Science Foundation under awards 2111688 and 2135579, and the Dwight D. Eisenhower Fellowship program under Grant No. 693JJ32345023. George Gunter was supported through an NSF Graduate Research Fellowship.}
}

\author{ MATTHEW W.~NICE\textsuperscript{\textdaggerdbl},
    GEORGE GUNTER\textsuperscript{\textdaggerdbl},
    JUNYI JI\textsuperscript{\textdaggerdbl},
    YUHANG ZHANG\textsuperscript{\textdaggerdbl},
    MATTHEW BUNTING\textsuperscript{\textdaggerdbl},\\
    WILL BARBOUR\textsuperscript{\textdaggerdbl},
    JONATHAN SPRINKLE\textsuperscript{\textdaggerdbl},
    AND
    DANIEL B.~WORK\textsuperscript{\textdaggerdbl}
    \thanks{\textdaggerdbl: Vanderbilt University}\
}

\maketitle

\begin{abstract}
This paper introduces a novel approach that seeks a middle ground for traffic control in multi-lane congestion, where prevailing traffic speeds are too fast, and speed recommendations designed to dampen traffic waves are too slow.  Advanced controllers that modify the speed of an automated car for wave-dampening, eco-driving, or other goals, typically are designed with forward collision safety in mind.  Our approach goes further, by considering how dangerous it can be for a controller to drive so slowly relative to prevailing traffic that it creates a significant issue for safety and comfort. This paper explores open-road scenarios where large gaps between prevailing speeds and desired speeds can exist, specifically when infrastructure-based variable speed limit systems are not strictly followed at all times by other drivers. Our designed, implemented, and deployed algorithm is able to follow variable speed limits when others also follow it, avoid collisions with vehicles ahead, and adapt to prevailing traffic when other motorists are traveling well above the posted speeds. The key is to reject unsafe speed recommendations from infrastructure-based traffic smoothing systems, based on real-time local traffic conditions observed by the vehicle under control. This solution is implemented and deployed on two control vehicles in heavy multi-lane highway congestion. The results include analysis from system design, and field tests that validate the system's performance using an existing Variable Speed Limit system as the external source for speed recommendations, and the on-board sensors of a stock Toyota Rav4 for inputs that estimate the prevailing speed of traffic around the vehicle under control. 
\end{abstract}


\section{Introduction}

Infrastructure-based freeway traffic control technologies are deployed on critical roadways to improve safety and mobility. Traditional systems include ramp metering to manage merging traffic onto the mainline, variable speed limit (VSL) systems that promote speed harmonization and reduce sudden slow-downs~\cite{lu2014review,sergeVSL,papageorgiou2008effects}, and lane control systems that provide information about lane-closures ahead due to crashes. Recently, the widespread commercial deployment of level 1 and level 2 automated vehicles has opened new opportunities for freeway traffic control, for example to stabilize the overall flow when only a small fraction of vehicles are equipped~\cite{ma2016freeway,stern2018dissipation,delle2022new}. Yet, today, the commercially available vehicle-based automation systems operate without coordination or cooperation with the infrastructure-based systems.

\begin{figure}
    \includegraphics[width=\linewidth]{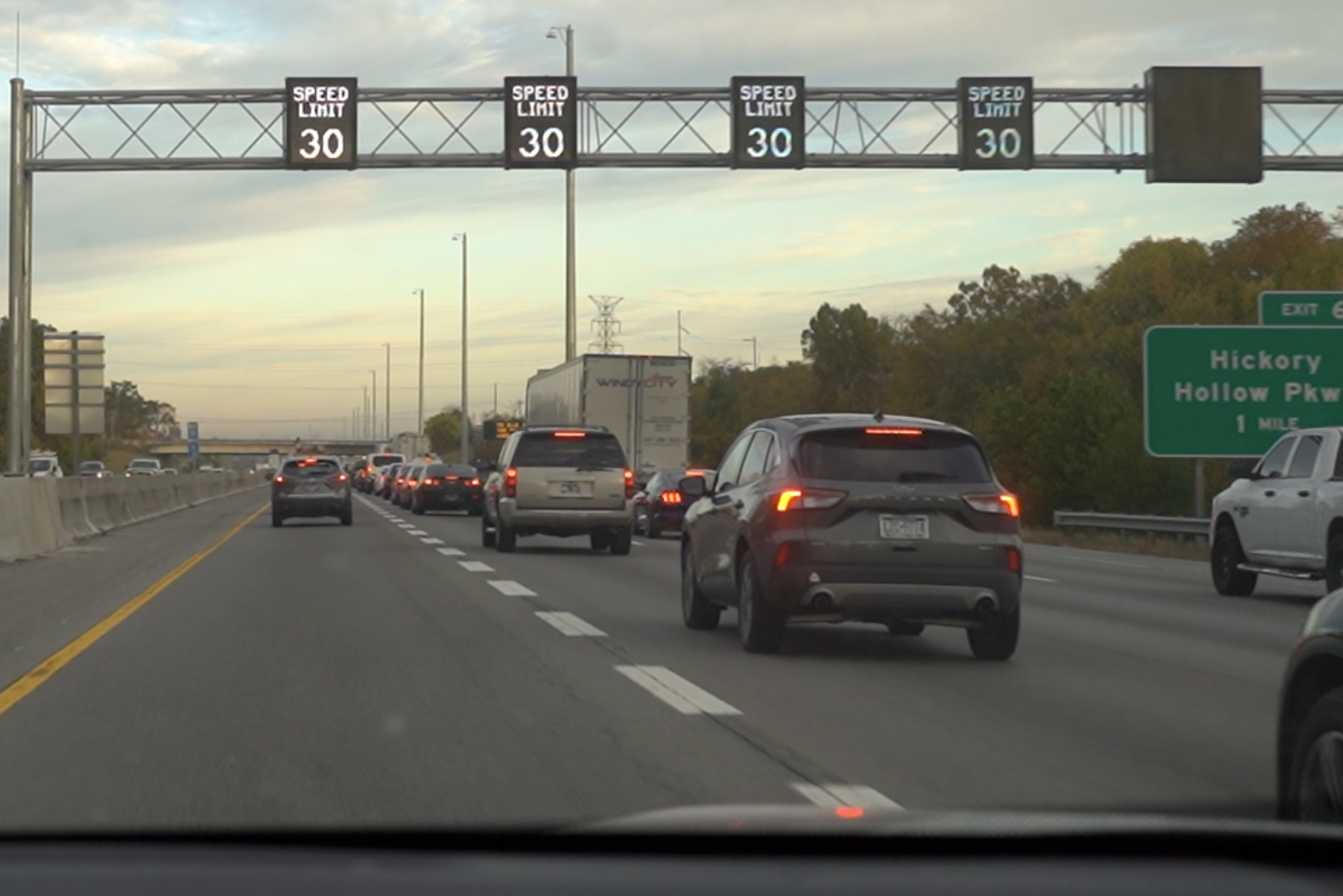}
    \caption{\textbf{Recurring Dilemma}: Variable Speed Limit (VSL) gantry shows a 30 mph speed limit on Interstate-24. Prevailing traffic is shown to regularly exceed the VSL by a large margin. In this work we demonstrate an automated vehicle controller that follows the VSL on the gantry when nearby vehicles do, and adopts a higher speed when prevailing traffic is moving much faster than the posted speed.}
    \label{fig:fall_vsl}
\end{figure}

In this paper, 
we consider the setting of cooperative variable speed limit control, in which  connected and automated vehicles (CAVs) adjust their speed to follow the infrastructure based variable speed limit  (VSL) system~\cite{grumert2015analysis}. In fully automated traffic flows, the problem of collaborative control is purely technical, e.g., designing the sensing, communication, and control systems to enable vehicles to follow the posted speed limits. However, in mixed autonomy settings, a pressing safety challenge arises from the inherent disparity between vehicles programmed to strictly follow speed limits and human-driven vehicles that frequently exceed these limits. As a motivating example, we have recently observed prevailing traffic as much as 30 mph above the posted variable speed limit on a major US freeway shown in Figure \ref{fig:fall_vsl}. Large gaps between the speed of traffic and the posted speed limit occur regularly in daily traffic jams. Naïve automated control of the vehicle to follow the speed limit rather than synchronizing vehicle speeds with the prevailing traffic flow will create unsafe conditions to unexpecting vehicles under human control. Simply following the prevailing traffic flow ignores the opportunity with CAVs to increase safety and efficiency on roadways.

Here is the main problem addressed in this work: \textit{How can we design a controller to follow variable speed limits when it can, while keeping up with the prevailing speeds when it needs to?}

We reason that automated vehicles must not drive substantially slower than human piloted vehicles if they are to be considered safe to operate in traffic and socially acceptable (and thus turned on, an obvious liveness constraint) by the owners of the equipped vehicles. This requirement to drive relative to the surrounding traffic creates new design challenges, given the timescales on which the traffic conditions change, and the inherent systematic latencies by many of today's commercial traffic information providers. These traffic state estimates provide updates on traffic conditions that are averaged in time and space, and have latencies in excess of a minute or more.

The main contribution of this work is to design, implement, and field test a new cooperative automated vehicle control algorithm that complies with variable speed limits when other human drivers do, and blends in with human drivers when they violate the posted speeds. The specific contributions are:

\begin{itemize}
  \item Introduction of a new notion of safety for cooperative automated vehicle applications to avoid causing controlled vehicles to drive substantially slower than surrounding traffic. Our approach recognizes the necessity for automated vehicles to adhere with the typical driving behavior observed on the roads, even if it requires a deviation from the posted speed limit. The control algorithm on the vehicle maintains collision avoidance through the use of a control barrier function-based safety filter, follows the posted speed limit when prevailing traffic is also operating near the speed limit, and exceeds the limit when prevailing traffic requires it to.  
  \item Development of a vehicular-based method for measuring prevailing traffic. Specifically we decode Controller Area Network (CAN) messages on a commercially available level 2 vehicle corresponding to the onboard radar unit, and use the observed radar measurements to estimate the speed of nearby downstream vehicles. Since the measurement is done on the vehicle, we can maintain safety (accurate awareness of with surrounding traffic) locally, even if we lose communication to external data sources. 
  \item Field experiments on two control vehicles operating in heavy morning rush hour traffic on the I-24 Freeway near Nashville, TN. We implement our controllers using low-cost hardware, 
  to enable 
  scalability of our approach. Our findings from the experiments show that we spend $16.6\%$ of time following the variable speed limit, $24.0\%$ of time above the speed limit due to prevailing traffic, and $59.4\%$ of the time in a  car-following mode to prevent forward collision.
\end{itemize}

The remainder of this article is organized as follows. In Section~\ref{sec:related_works} we review the most closely related works on infrastructure-based and vehicle-based traffic control. In Section~\ref{sec:methods} we describe our control system. In Section~\ref{sec:experiment_setup} we review the experimental setup. Section~\ref{sec:results} provides the findings from the field test of our controller operating in heavy traffic.

\section{Related Works}
\label{sec:related_works}

\subsection{Variable Speed Limit Systems}
Since the initial deployment of VSL around 1960s, the effectiveness of VSL on road safety and mobility has been investigated in both simulation and field tests~\cite{lu2014review, khondaker2015variable, zhang2023marvel}. Empirical studies have reported important safety findings. For example, a Belgian study reported an 18\% reduction in injury crashes and a 20\% decrease in rear-end collisions after the VSL implementation~\cite{de2018safety}. 
Similarly, a study conducted over 72 months in Seattle demonstrated a 32.23\% reduction in overall crashes, with the most significant impact observed in rear-end collisions~\cite{pu2020full}.

While the safety benefits of VSL are generally positive, their effectiveness is sensitive to the rate of driver  compliance~\cite{hadiuzzaman2015modeling, hellinga2011impact}. Simulation studies affirm that the compliance rate is crucial for the performance of VSL~\cite{habtemichael2013safety, zhang2022quantifying}. Challenges in implementing automated speed enforcement in North America contribute to low compliance \cite{hellinga2011impact}. Additionally, the large gap between posted and prevailing speeds can further impair driver compliance~\cite{nissan2011evaluation}.

\begin{figure*}[!ht]
    \centering
    \includegraphics[width=\linewidth]{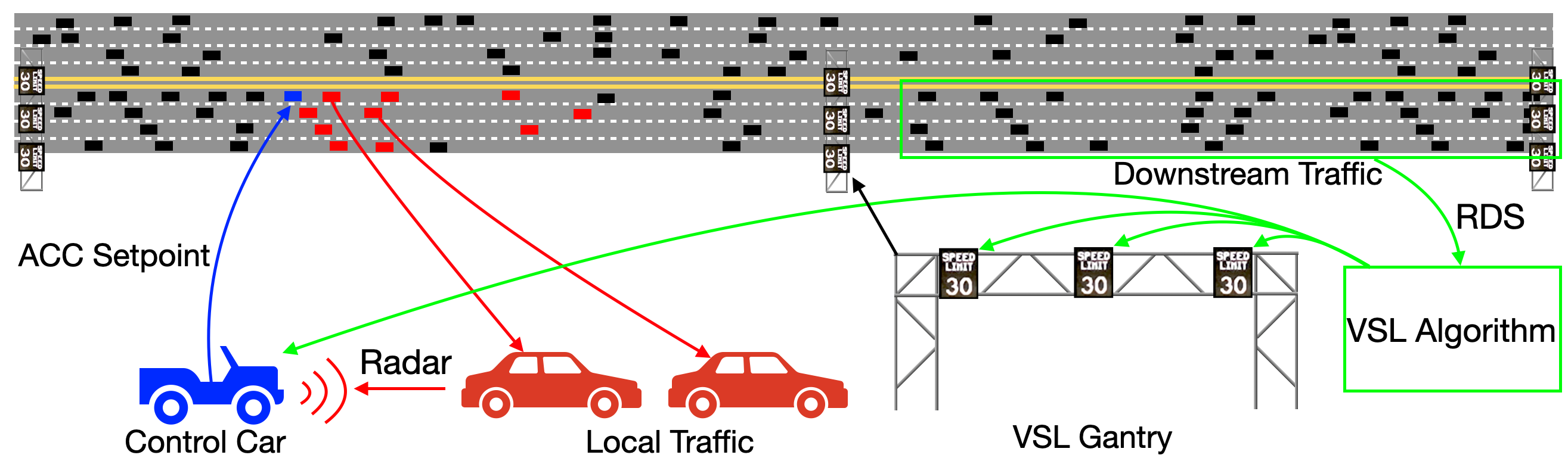}
    \caption{\textbf{Environment}: Overview of the control environment.  The VSL system measures downstream traffic for aggregate traffic information.  A control vehicle (blue) can measure timely information about highly local traffic in front and adjacent to the vehicle (red).  Our controller changes the set speed based on information from both of these sources.}
    \label{fig:cps}
\end{figure*}

\subsection{Connected and Automated Vehicles}
In recent years, CAVs have been considered as mobile actuators in the traffic stream~\cite{stern2018dissipation,delle2022new}. It has been shown that CAVs have the potential to reduce congestion and decrease the total travel time~\cite{talebpour2016influence}. In line with the concept of the living laboratory~\cite{lochrane2014using}, CAVs have been implemented at a scale of 100\cite{nice2023enabling} to provide a distributed testbed for novel control with the open freeway as the living lab. Experimentation in real-world settings is a key to cyber-physical systems research where the physical systems are often complex, and interactions are nuanced, such as in transportation systems. In the realm of control strategies, Xiao et al.~\cite{xiao2021rule} developed a framework to design optimal control strategies for automated vehicles  that are required to satisfy a set of traffic rules with a given priority structure. Schwarting et al.~\cite{schwarting2019social}  discussed how CAVs can adapt to social preferences of other drivers by integrating social psychology tools into controller design, thereby improving autonomous performance. Nice et al.~\cite{nice2021can} explored the role of human-in-the-loop cyber-physical systems in traffic flow control and introduced the ``CAN Coach", a system that augments human perception with radar data to improve vehicle control.

\subsection{Vehicle to Everything}
Vehicle-to-everything (V2X) is an overarching term that encompasses various forms of vehicle communication, including vehicle-to-vehicle (V2V) and vehicle-to-infrastructure (V2I). 

V2V communication is pivotal for the future of intelligent transportation systems, particularly for CAVs. V2V has been applied to reduce time headway for platooning of connected vehicles, thereby enhancing traffic flow~\cite{bian2019reducing}. In addition, \cite{aoki2018dynamic} presents a cooperative dynamic intersection protocol for CAVs, utilizing V2V communications and perception systems, to safely and efficiently navigate these intersections. The proposed protocol significantly improves traffic throughput and minimizes trip delays when compared to baseline models. 

Complementing V2V, V2I focuses on the interaction between vehicles and road infrastructure. There is a growing body of simulation studies that explores the integration of VSL and CAV within the broader context of V2I communication. For instance, Li et al.~\cite{li2017integrated} demonstrated that integrating V2I with VSL and cooperative adaptive cruise control (CACC) can effectively reduce rear-end collision risks. Furthermore, Grumert et al.~\cite{grumert2015analysis} showed that the benefits of V2I communication, autonomous vehicle control, and individualized speed limits for VSL systems result in harmonized traffic flow and reduced exhaust emissions. 

There are few studies of V2I field experiments mainly because the lack of widely available CAVs and communication gaps between the infrastructure operators and the vehicle automation systems. Ma et al.~\cite{ma2016freeway} conducted a field experiment on an active freeway with recurring congestion, employing three V2I-equipped vehicles to implement a simple speed recommendation algorithm. The study used probe vehicles to measure the impacts on the overall traffic flow and found that the V2I-enabled speed recommendation algorithm reduced oscillatory behavior without negatively affecting travel times. The control effectiveness from a small portion of automated vehicles has been further demonstrated in~\cite{zhang2022quantifying}, which shows that a small number of vehicles complying with the speed limit has a greater \textit{effective compliance rate} since non-complying vehicles have limited ability to maneuver around complying ones. A recent study~\cite{nice2023sailing} developed a modified vehicle controller that is able to dynamically adjust the speed according to the posted speed limits by using LTE connectivity. They deployed the controller on a congested highway segment and found that the control vehicle resulted in a 25\% reduction of speed variability compared to a human-driven probe vehicle in the same traffic stream.

\section{Methods}\label{sec:methods}

Our design challenge is to architect a vehicle speed controller that considers both the legally enforceable variable speed limit, and local traffic relative to the controlled vehicle.  Figure~\ref{fig:cps} outlines the environment of the design involving a single control vehicle and two sources of traffic information.  The controller acts as a replacement for the OEM Adaptive Cruise Control (ACC).  The design integrates both downstream and local traffic conditions to switch into 5 different control modes:

\begin{itemize}
    \item \textit{Normal-Mode}: When not on a roadway with a VSL system and no vehicle is in front, then operate like a standard cruise controller.
    \item \textit{VSL-Mode}: Drive at the speed setpoint provided by the VSL. This can occur if there is no traffic, or we do not meet the conditions to enter the other modes. This is a V2I interaction that is only be engaged while operating on a roadway with a VSL system. 
    \item \textit{Middleway-Mode}: If nearby traffic is driving much faster than the variable speed limit, then control the vehicle speed at a middle ground between the VSL speed and prevailing traffic. This is effectively driving in a reduced go-with-the-flow behavior. 
    \item \textit{CBF-Mode}: If a lead vehicle is in front and driving slower than the current speed setpoint then follow the leader in manner which will prevent collisions using a control barrier function (CBF). 
    This mode overrides the other active modes at any time. This is similar to a stock ACC system.
    \item \textit{Disengaged}: Control is inactive, driver has full control.
\end{itemize}

Normal-Mode is mutually exclusive to VSL-Mode and Middleway-Mode.  When the vehicle is not on a roadway with a VSL system, the controller will only use the modes of the Normal-Mode or CBF-Mode to mimic the OEM ACC. 

We will first describe the design of the controllers to acheive these modes, then describe the specific implementation including the location, vehicles, and vehicle hardware.

\subsection{Controllers}



Here we describe the different controllers and mode switching on the experimental vehicle throughout testing.  Downstream traffic information is provided by the VSL system, which is primarily responsible for setting the variable speed limit on the gantries.  Local traffic information is measured through onboard sensors on the car, such as radar.  Information is fused from both sources to provide a speed setting for the controller vehicle.  The design of the custom cruise controller is based on a hierarchy of a low-level speed and safety controller in tandem with higher-level speed setpoint selection algorithm.  We start by looking at the design of the higher-level speed selection, then described the lower-level speed control.

\subsubsection{Speed Selection}

\begin{figure}
    \centering
    \includegraphics[width=\linewidth]{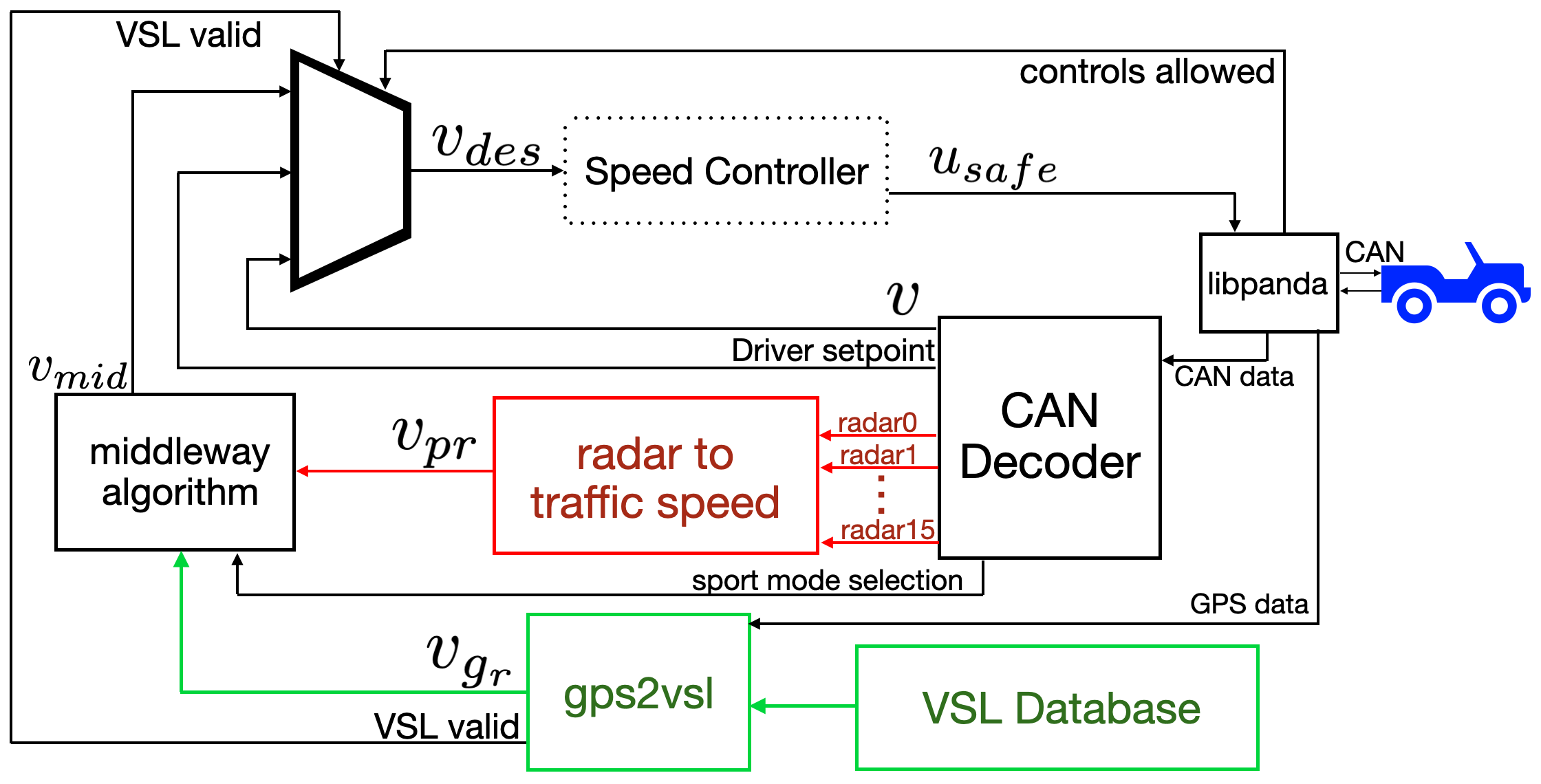}
    \caption{\textbf{Speed Selection}: The system architecture to determine the speed setting based on the VSL gantry, the state of the cruise controller, GPS information, and the radar data.  The equation for the middleway algorithm is shown in equation~\ref{equ:middleway}.}
    \label{fig:speedsetting}
\end{figure}


Figure~\ref{fig:speedsetting} shows the design of selecting a speed setting for the speed controller.  The speed setpoint is switched between three different sources through a multiplexer.  The speed setting is either set to the current speed of the vehicle $v$ (Disengaged), the driver's setpoint as set by the cruise controller interface, or the middleway algorithm that uses local and downstream traffic information $v_{mid}$.  The setpoint is chosen based on the state of the vehicle and the location and direction of the vehicle.  

\begin{itemize}
    \item If the driver has not yet engaged the cruise controller then controls are not allowed (Disengaged), so the multiplexer sends the vehicle's current speed $v$ as the controller setpoint.  This is done to ensure that a smooth transition occurs when the driver engages the controller.
    \item If the controller is engaged but the vehicle is not in the VSL environment, then the multiplexer will switch the setpoint to the driver's setting from the gauge cluster (Normal-Mode/CBF-Mode). 
    \item If the controller is engaged, vehicle is within the VSL region, and has a valid VSL reading, then the multiplexer switches to the setpoint provided by the middleway algorithm $v_{mid}$ (VSL-Mode/Middleway-Mode/CBF-Mode).
\end{itemize}

The middleway algorithm shown in Figure~\ref{fig:speedsetting} is defined as follows. Let $v_{mid}$ be the output desired speed of the middleway algorithm, $v_{g_{r}}$ be the recommended speed from the relevant gantry in the VSL system, $v_{pr}$ be the average velocity of faster moving vehicles observed by the control vehicle's forward radar sensor, and $v_{des_{max}}$ be the highest allowable $v_{mid}$.  
\veloffset
is a runtime threshold parameter representing the how much slower than $v_{pr}$ the vehicle operator is comfortable with, and is settable by the driver through the vehicle's \emph{Sport Mode} and \emph{Eco Mode} features.  The desired speed is then calculated using a control law of the following form:

\begin{equation}
\label{equ:middleway}
    v_{mid} = \min(\max(v_{pr} - \veloffset, v_{g_{r}}), v_{des_{max}})
\end{equation}

Note that the usage of the max function in \eqref{equ:middleway} effectively encodes the mode switch between Middleway-Mode and VSL-Mode.

The local speed of traffic, $v_{pr}$ is estimated when there are vehicles going faster than the ego vehicle. The estimate takes a point cloud of radar measurements from the last 5 seconds, and averages the observations from vehicles going faster. If there are not enough recent observations, the estimate is switched off (outputs $0$) and the $v_{mid}$ is consequently the posted VSL speed $v_{g_{r}}$.


\subsubsection{Speed Controller}
\begin{figure}
    \centering
    \includegraphics[width=\linewidth]{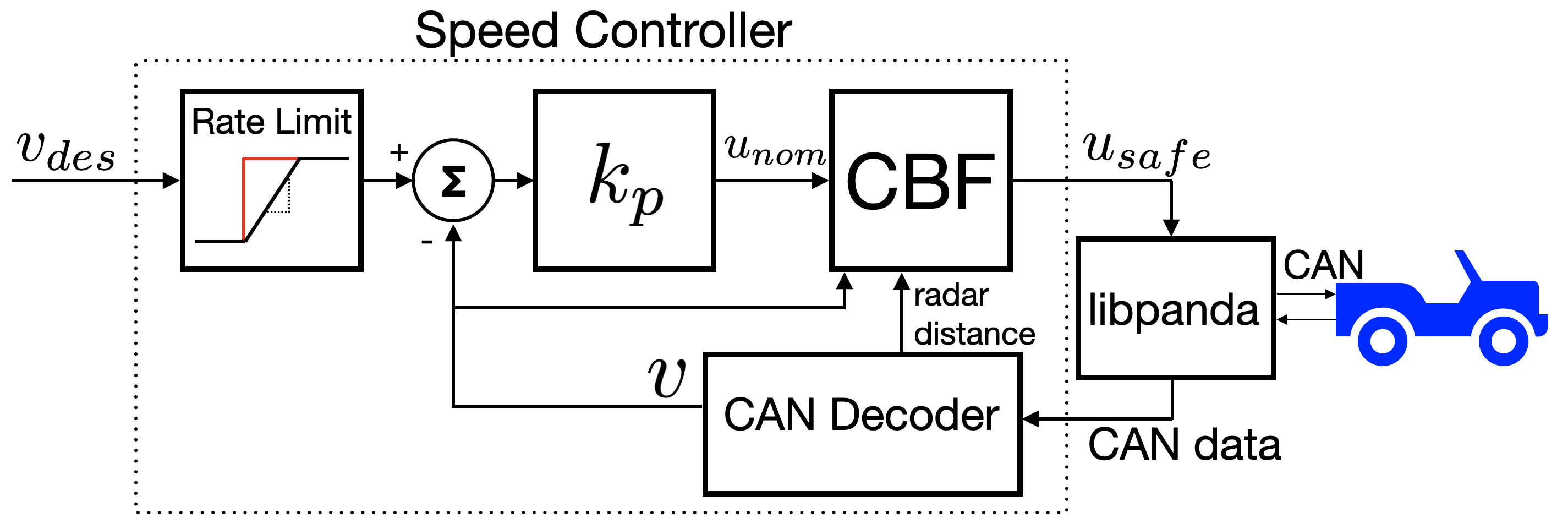}
    \caption{\textbf{Speed Controller}: this acceleration-based controller is a replacement for the OEM cruise controller.  The controller takes an input speed setting, $v_{des}$, and sends acceleration commands to the vehicle through libpanda.  The speed controller is based on rate limiting $v_{des}$, a nominal proportional controller, and a CBF to perform dynamic filtering to provide car following and prevent collisions.}
    \label{fig:speedcontrol}
\end{figure}

Figure~\ref{fig:speedcontrol} shows the low-level controller design.  Vehicular dynamics are controlled via a commanded acceleration value sent along the vehicle's CAN bus using libpanda\cite{bunting2021libpanda}. A low-level control system implemented by the vehicular manufacturer converts this command to more specific vehicular dynamic commands (e.g. throttle, braking, engine).

First, a time-based ramp function is applied to the $v_{des}$ input of the nominal controller to produce $v_{ramp}$.  This was designed for use cases when the setpoint may exhibit discrete jumps.  Using a ramp function rate-limits the input and allows the setpoint to be changed without potentially unsafe transient effects feeding through to actuation.  In our specific use case of dynamically changing the setpoint during the experiment the ramp function allows for switching between setpoints from different sources.

The vehicle acceleration request $u$ is based on two control algorithms. The first is the control law we refer to as the \textit{nominal controller}, which calculates acceleration commands meant to track the filtered desired speed $v_{ramp}$. Let $u_{nom}$ refer to the acceleration coming from the nominal controller. 

The acceleration from the nominal controller is calculated using a proportional control law of the following form:

\begin{equation}
    u_{nom} = k_{p}\left( v_{ramp} - v \right)
\end{equation}
where $k_p$ is the proportional gain parameter, $v_{ramp}$ is the rate-limited desired velocity, and $v$ is the instantaneous velocity. in our specific implementation, a value of $0.8$ was used for $k_p$ in experimentation. 

The CBF-Mode employs a low-level supervisory controller based on a control barrier function that overrides engaged controllers to avoid forward collisions. For a formal description of the design of this CBF, see~\cite{gunter2022experimental,gunter2022CBFs_for_cutins,ames2016control}. The form of the controller is as follows:
\begin{equation}
    u_{\mathit{safe}} = \frac{k_{\mathit{CBF}}}{t_{min}}\left(s - \left( t_{min}v + s_{min} \right) \right) + \frac{1}{t_{min}} \left( v_l - v \right)
\end{equation}
where $u_{\mathit{safe}}$ is the maximum allowable safe control acceleration, $s$ the inter-vehicle spacing, and let $v_{l}$ the speed of the lead vehicle immediately ahead. $k_{\mathit{CBF}},t_{min},\text{and } s_{min}$ are control parameters which we assign values of $0.1, 2.0,$ and $15.0$ respectively. This CBF is designed to filter control accelerations so that the vehicle's spacing-gap stays above a value of $t_{min}v + s_{min}$. This choice of safety is common~\cite{ames2019control,ames2016control}, but not unique. 
For example, this safety choice and design has been safely and effectively fielded in other open-road field tests, such as \cite{nice2023sailing}.

\subsubsection{Controller behavior at scale}
Even though our controller allows travelling above the posted VSL, if a series of vehicles run it, the traffic flow will approach VSL speeds or slower. Middleway-Mode is only needed as long as enough of the traffic flow continues to violate the speed limit, creating the scenario where following the law and maintaining safety by matching traffic flow conflict. 

Consider a highway where all travelling vehicles are control vehicles where the penetration rate $p=1$. There are two cases where traffic is not following the VSL speed $v_{g_{r}}$: either traffic is faster than $v_{g_{r}}$, or traffic is slower than $v_{g_{r}}$. 

In the case where traffic is faster, consider a vehicle $n$ that observes vehicles downstream of itself moving faster such that $v_{pr}^{n}-\veloffset^{n} > v_{g_{r}}$. Vehicle $n$ would travel tracking some speed $v_{pr}^{n}-\veloffset^{n}$ slower than $v_{pr}^{n}$. Consequently, the vehicle upstream of $n$, vehicle $n+1$, would observe some $v_{pr+1}^{n+1} < v_{pr}^{n}$ and travel slightly slower than $n$. A series of control vehicles will then eventually approach $v_{g_{r}}$. For each vehicle $m$ running our controller, when $v_{pr}^{m}-\veloffset^{m} = v_{g_{r}}$, the controller's $v_{des}^{m}$ will start to track $v_{g_{r}}$ directly, i.e. VSL-Mode. 

In the case where traffic is slower than $v_{g_{r}}$, the desired velocity $v_{des}^{m}$ for all control cars $m$ would still stay at $v_{g_{r}}$, however the CBF-Mode is empowered keep vehicle speeds slower than $v_{g_{r}}$ to maintain forward safety. 

In this work, the penetration rate $p\approx0$, however this control scheme is suitable to be used as $p$ increases in possible future deployments.

\subsection{Hardware and Software Implementation}

The prior section described the controller design, agnostic to specific implementation.  This section discusses specific implementation in the environment, the vehicle control implementation, the vehicle-to-infrastructure interface, and the control vehicle computing hardware instrumentation.

\subsubsection{Vehicle System Architecture}
\begin{figure}
    \centering
    \includegraphics[width=\linewidth]{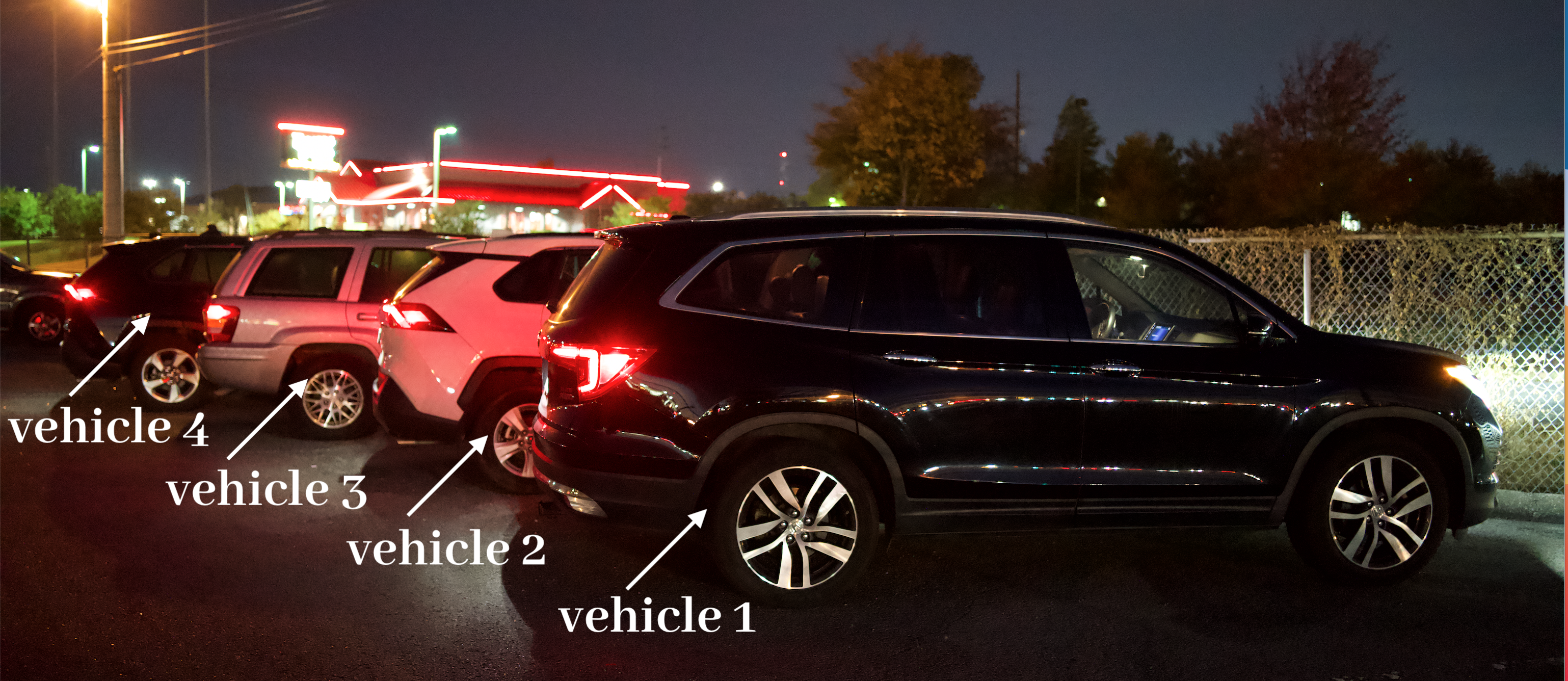}
    \caption{\textbf{Experimental Deployment}: Four vehicles, pictured here, are launched into early morning congestion on Interstate-24. From right to left, they enter into the traffic flow. Vehicles 2 and 4 are instrumented for experimental control, and vehicles 1 and 3 are operated under human-piloted control.}
    \label{fig:4cars}
\end{figure}

The control system was installed on two different Toyota Rav4s.  In conjunction with the control cars, two additional cars were equipped with GPS recorders (Figure~\ref{fig:4cars}). Using low cost hardware, the vehicle system accesses the VSL data through a web-based pipeline over an LTE connection with a tethered mobile phone. Vehicle control commands are created by combining the vehicle-to-infrastructure connectivity with the vehicle's proprioception, and the vehicle's local traffic state exteroception. We leverage the Robotics Operating System (ROS) message framework \cite{quigley2009ros} for system integration.  The structure in Figures~\ref{fig:speedsetting} and \ref{fig:speedcontrol} are designed as ROS nodes and topics.  Our system can be broken down into three categories: vehicle interfacing, VSL integration, and control design.


\subsubsection{Hardware Instrumentation}
Each of the 4 vehicles were equipped with the same set of hardware for both use cases of data collection and vehicle control.  This includes a Raspberry Pi 4 running Raspbian and ROS.  A USB GPS module based on the uBlox m8 provided location and time information.  A board called the mattHat provided the CAN interface provided CAN reading in all 4 vehicles, and control in the 2 control vehicles.  For live VSL database connectivity to get the latest setpoints, mobile phones provided a hotspot over USB cables using a utility called usbmux.  Excluding the mobile phones, each hardware kit cost less than \$500 USD.


Interfacing with the control vehicle was performed using the software libpanda\cite{bunting2021libpanda}.  Libpanda has the capability to firewall CAN messages between system modules designed by the \textit{original equipment manufacturer} (OEM), allowing third party messages to replace OEM messages. Through libpanda, both on-board measurements can be read/recorded, and control commands can be sent to the vehicle. In tandem with CAN interfacing, libpanda also interfaces with USB GPS modules to provide position information. Libpanda also keeps track of the OEM Advanced Driver Assistance System (ADAS) module state to prevent hardware-level errors when attempting to engage the driving automation system.  

Code generation techniques as in \cite{nice2023middleware} are used to convert manufacturer-specific vehicle CAN message into a homogenenous framework in ROS.  The vehicle interface is a ROS node with an autogenerated CAN parser that produces sensor data like radar signals and cruise control setpoint. The vehicle interface node is a part of the can\_to\_ros project \cite{elmadani2021can}, exposing CAN-level vehicle systems to ROS. The radar sensor is among the CAN-level sensors.  In the case of both of the controlled Toyota Rav4s, the radar produces up to 16 tracks of point cloud data along with relative speed at each point.

\subsubsection{Active Traffic Management Infrastructure}
The experiment is held on a section of Interstate-24, specifically in the I-24 SMART Corridor~\cite{tdot2023i24} located near Nashville, Tennessee.  This section is part of an \textit{active traffic management system} (ATMS) to improve safety and reliability.  VSL gantries are installed approximately every 0.5 miles to provide speed limits for all lanes, which can change at 30 second intervals.  The posted speed limits can vary from 30 mph to 70 mph.  A ROS node named gps2vsl can access the information posted to each VSL gantry from a basic URL request, discussed further in Section \ref{sub:gps2vsl}.

\subsubsection{Data Integration from Public Infrastructure}

Messages are sent from the traffic operations center to each gantry whenever the speed limit should change. Our VSL data feed is obtained from a database mirror that records these messages from the traffic operations center to the gantries.

\subsubsection{Models for gps2vsl}
\label{sub:gps2vsl}
The gps2vsl node uses the latest position information provided by libpanda, and compares its location against the static set of VSL gantry positions.  Geofencing is used to first check that the vehicle is within the I-24 SMART corridor, otherwise a ROS topic informs that the current VSL setpoint is invalid.  Once inside the corridor, GPS is used to approximate the vehicle's heading to select either the east bound or west bound gantries.  With direction known, the car location is compared against the locations of the gantries.  If the vehicle enters within 0.15 miles of a gantry, then that specific gantry's speed limit will be published as a setpoint, along with informing other ROS nodes that the VSL setpoint is valid. Checking the pertinent variable speed limit for the vehicle occurs when entering the bounds of the downstream gantry and every 5 seconds, in order to capture VSL changes by location and over time.

\section{Experimental Setup}
\label{sec:experiment_setup}
The experimental control vehicle deployment consists of four vehicles. There are two pairs of vehicles; each pair has one control vehicle and a preceding `probe' vehicle recording trajectory data. The speed of the first vehicle of each pair is regulated by the driver, who is instructed to maintain a safe driving speed at all times. Practically this results in drivers traveling close to the prevailing traffic speed. The speed of the second vehicle in each pair, or `control vehicle', is regulated by the novel control system introduced in this work. All drivers were instructed to maintain a safe operating environment and to abandon the experiment if conditions on the roadway prevent a safe experiment from being executed. The experiments were conducted in the 5:30-8:30 am window in which I-24 experiences the start of morning traffic conditions and regular traffic waves develop.

The experiment is conducted on a segment of I-24W with four lanes, which starts from the mile marker 70 and ends at mile marker 53. The vehicles are instructed to operate in the left-most lane on the roadway. The vehicles enter the roadway upstream of traffic waves, and then travel through the heavy congestion where the VSL activates and stop-and-go waves are observed. The current equipped VSL algorithm on SMART Corridor is designed to harmonize traffic speeds and is a modified version of the algorithm described in~\cite{zhang2022quantifying}. In particular, the VSL controller is activated when the observed traffic characteristics exceed predefined thresholds and the speed limits will be rounded to the nearest multiple of 5. 

\section{Results}\label{sec:results}

\begin{figure*}
    \centering
    \includegraphics[width=0.7\linewidth]{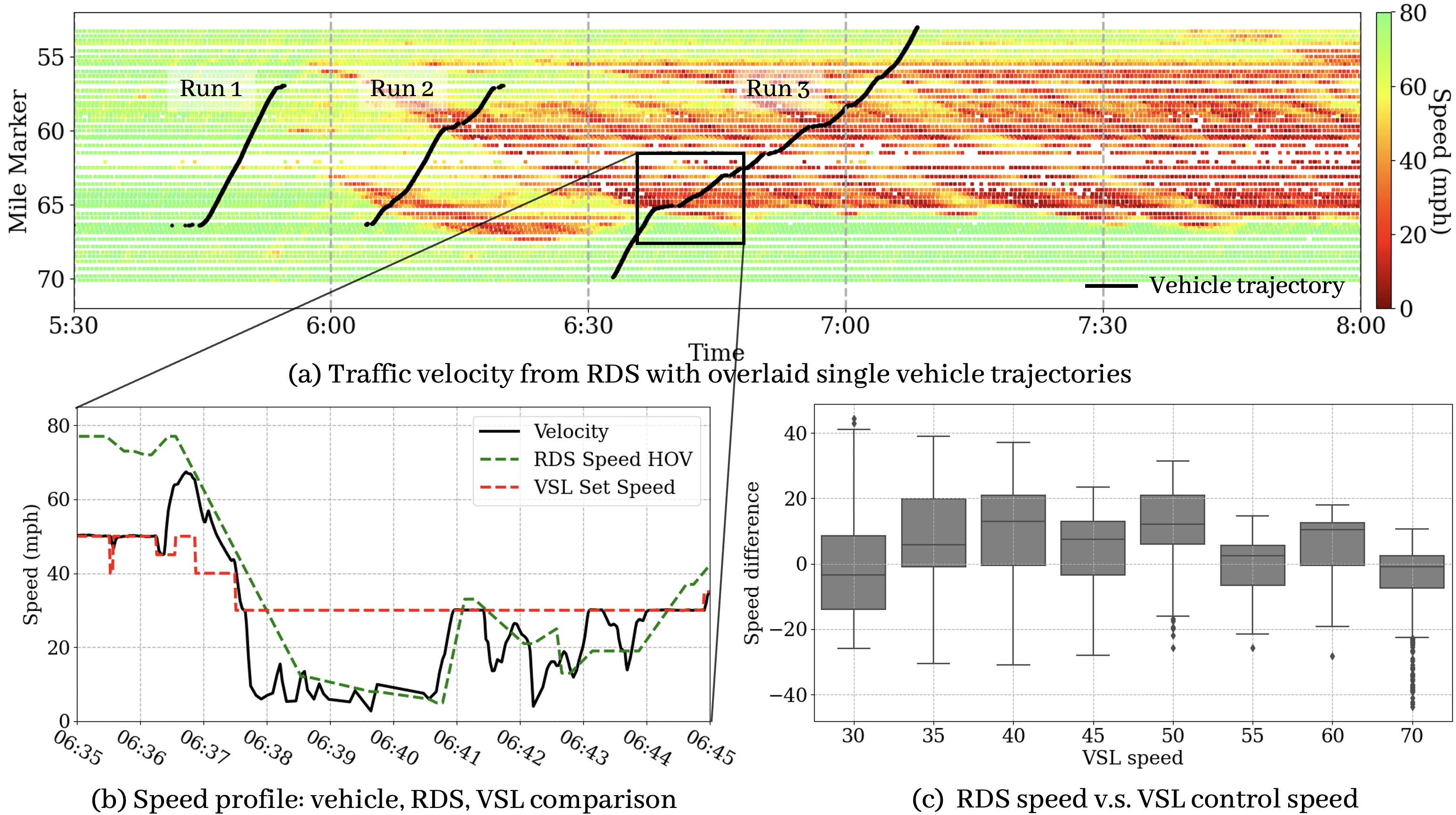}
    \caption{\textbf{Measuring the Discrepancy Between Prevailing Speed and VSL}: In three parts, this figure shows the context of the main problem posed in this work. (a) shows estimates of the traffic state from fixed-infrastructure Radar Detection System (RDS) sensors along the SMART Corridor on 08/29/23, with overlaid trajectories from a single vehicle making three Westbound trips. X-axis is time, and y-axis is the roadway mile markers, with the direction of travel going upward. Note the consistent green area below the wall of red; this is where congestion starts. Also note the recurring changes between red/orange/yellow; these are `stop-and-go' traffic waves. (b) shows the recurring dilemma an individual driver is faced with: when approaching a slowdown, either follow the posted speed limit, or keep up with traffic? In the minutes before a near stop, we observe a 25 mph+ discrepancy between the posted speed limit and the prevailing speed of traffic. This discrepancy resurfaces often, at the peak of traffic waves before the next stop. (c) expands the comparison of RDS (dotted green) and VSL (dotted red) in (b) to the the entire morning's traffic (05:00-09:59). The distribution of differences in speed show that the prevailing speeds regularly reach 10mph-20mph over the speed limit. $23.9\%$ of RDS-measured traffic speeds exceed 10mph over the variable speed limit during morning traffic.}
    \label{fig:vsl-challenge} 
\end{figure*}

This section summarizes the main findings of our implementation and experiments in real traffic. First, we will show that the prevailing traffic speed 
regularly  exceeds the posted variable speed limit (VSL) by 10 mph or more on the freeway of interest. This quantifies and validates our anecdotal observations that motivated our design. Next, we establish that local traffic estimates need to be minimally latent to be accurate enough to understand the local traffic state in real time, which supports our decision to measure traffic locally on the vehicle rather than to rely on external traffic sources. Finally we highlight the behavior of our controller in live traffic, showcasing that the various control modes are all regularly used when navigating complex freeway traffic. 

\subsection{Traffic Speed Far Exceeds Posted Speed Limits}\label{sec:motive}

The speed 
of traffic is regularly much faster than the posted speed limit (10 mph and higher). Figure~\ref{fig:vsl-challenge} shows this in three parts: the macroscopic traffic patterns, the perspective of an individual vehicle, and the trends in comparison between the infrastructure-based average speed observations and the posted variable speed limit at the time and place of observation.

The time-space diagram in Figure~\ref{fig:vsl-challenge}~(a) shows the typical onset of congestion on I-24 Westbound. This plot is a pairing of fixed infrastructure Radar Detection System (RDS) and an instrumented vehicle recording its trajectories.  With time in the x-axis, and the roadway direction going up the y-axis, vehicle trajectories (black) run up and to the right. Consequently, the slope of the trajectory is the velocity of the vehicle; a stopped vehicle creates a horizontal line. Around 6:00AM, traffic waves begin. Before 6:30, an approximately 10 mile region of congestion has formed and will continue for the next couple of hours. There is consistently a large sudden slowdown around MM 67, and a large number of traffic waves shown in alternating red and yellow regions. This overview conveys the typical congested traffic patterns on this roadway.

Figure~\ref{fig:vsl-challenge}(b) takes a closer look at this area of congestion from the perspective of the vehicle trajectories featured in Figure~\ref{fig:vsl-challenge}(a) in black. The variable speed limit is set between 50 mph and 40 mph in the region just upstream of the stopped traffic (06:35), giving an indication to all vehicles that they can anticipate a slow down. At the same time, the speed 
of traffic continues to travel at an average of of approximately 75 mph until just before 06:37. Before 06:38 the vehicle velocity (black) is below 10 mph and the RDS measuring aggregate speeds in this lane meets this observation at its next provided measurement (every 30s).

The traffic flow does not slow until a minute before meeting a wall of congestion and slowing to nearly a stop. This presents the operator of a control vehicle which only follows the posted speed limit two options: (1) follow the posted speed while the prevailing conditions are 25 mph+ higher, or (2) disengage the controller to support comfort and safety. This decision scenario repeats in the canonical traffic waves of the congestion region, every few minutes. The rest of the results section shows how we address this dilemma: a velocity controller which sees a middle way between the VSL and the high prevailing speeds. The control vehicle, being aware of the traffic 
speed
in local surroundings, of the variable speed limit setting on the roadway, and of the forward collision safety, has a new way to ride the traffic waves in morning congestion. It does this by compromising between traveling fast enough to be ride comfortably in prevailing traffic, while also supporting the pro-safety and wave dampening goals of the VSL's active traffic management system.


\subsection{Latency in Measuring Traffic Speed 
Induces Error}
\begin{figure}
    \centering
    \includegraphics[width=\linewidth]{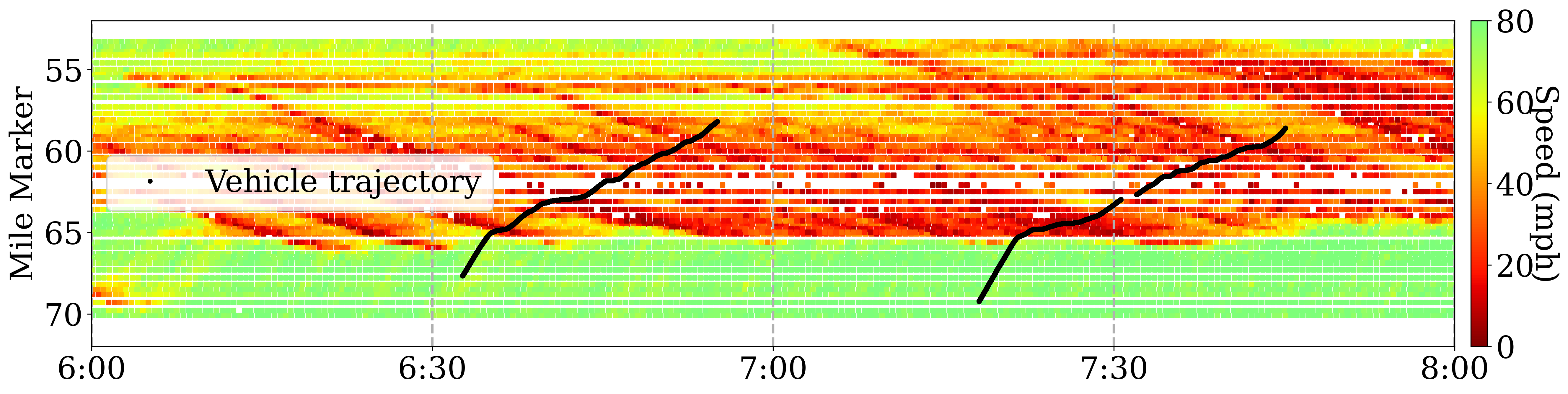}
    \caption{\textbf{Middle Way Control Deployed} Trajectories of a control vehicle are shown, laid over RDS speed measurements from the lane of travel. As in Figure~\ref{fig:vsl-challenge}, X-axis is time, and Y-axis is the roadway mile markers, with the direction of travel going upward.}
    \label{fig:spacetime}
\end{figure}

Here we describe the effect that latencies in fixed-infrastrucutre Radar Detection System (RDS) have on the accuracy of estimating prevailing traffic speeds. First, to create an estimate of prevailing speeds from RDS that we consider `ideal', we compare the trajectory of the test vehicle back to the historical speed measurements. Every point along the test vehicle's freeway trajectory is mapped to the 4 RDS speed measurements in space-time that contain that point. The ideal measure of the prevailing traffic speed at that trajectory point is then calculated by taking the average of these 4 points, which would not be possible in real-time. Figure~\ref{fig:spacetime} shows the RDS speed measurements captured in the test lane, as well as the control vehicle's trajectory. 

\begin{figure}
    \centering
    \includegraphics[width=\linewidth]{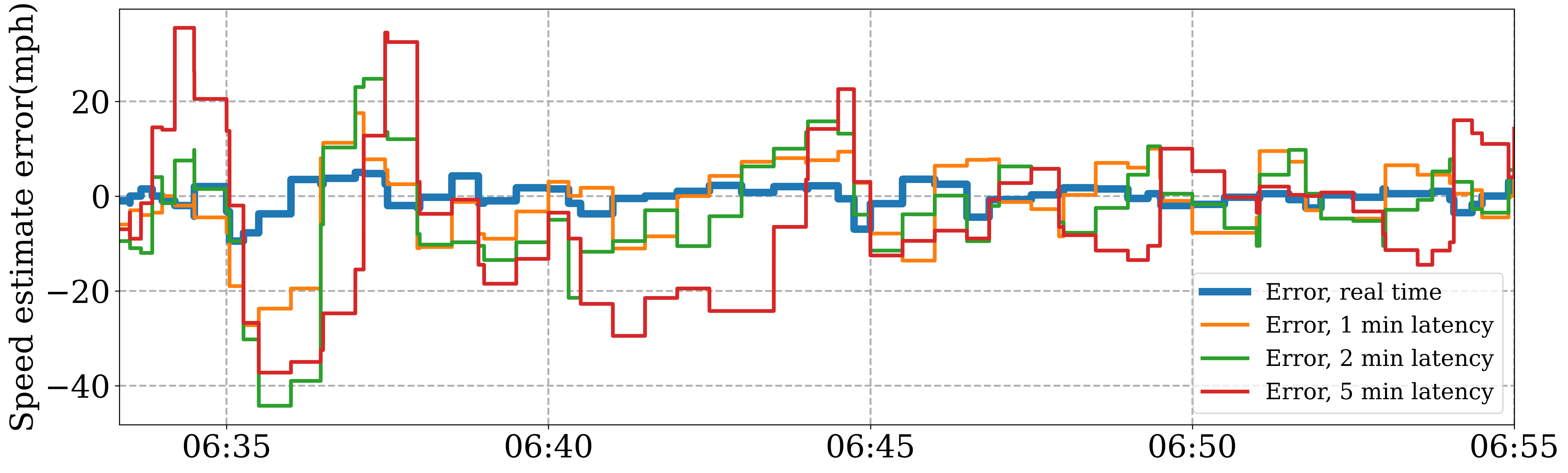}
    \caption{\textbf{Latency-Induced Errors}: Estimates of the speed of traffic are made, showing the effect of latency over time. Small errors in estimation are exacerbated with latency, because of how quickly the state of the traffic system changes in congested regions.}
    \label{fig:speed_error}
\end{figure}

\begin{figure}
    \centering
    \includegraphics[width=\linewidth]{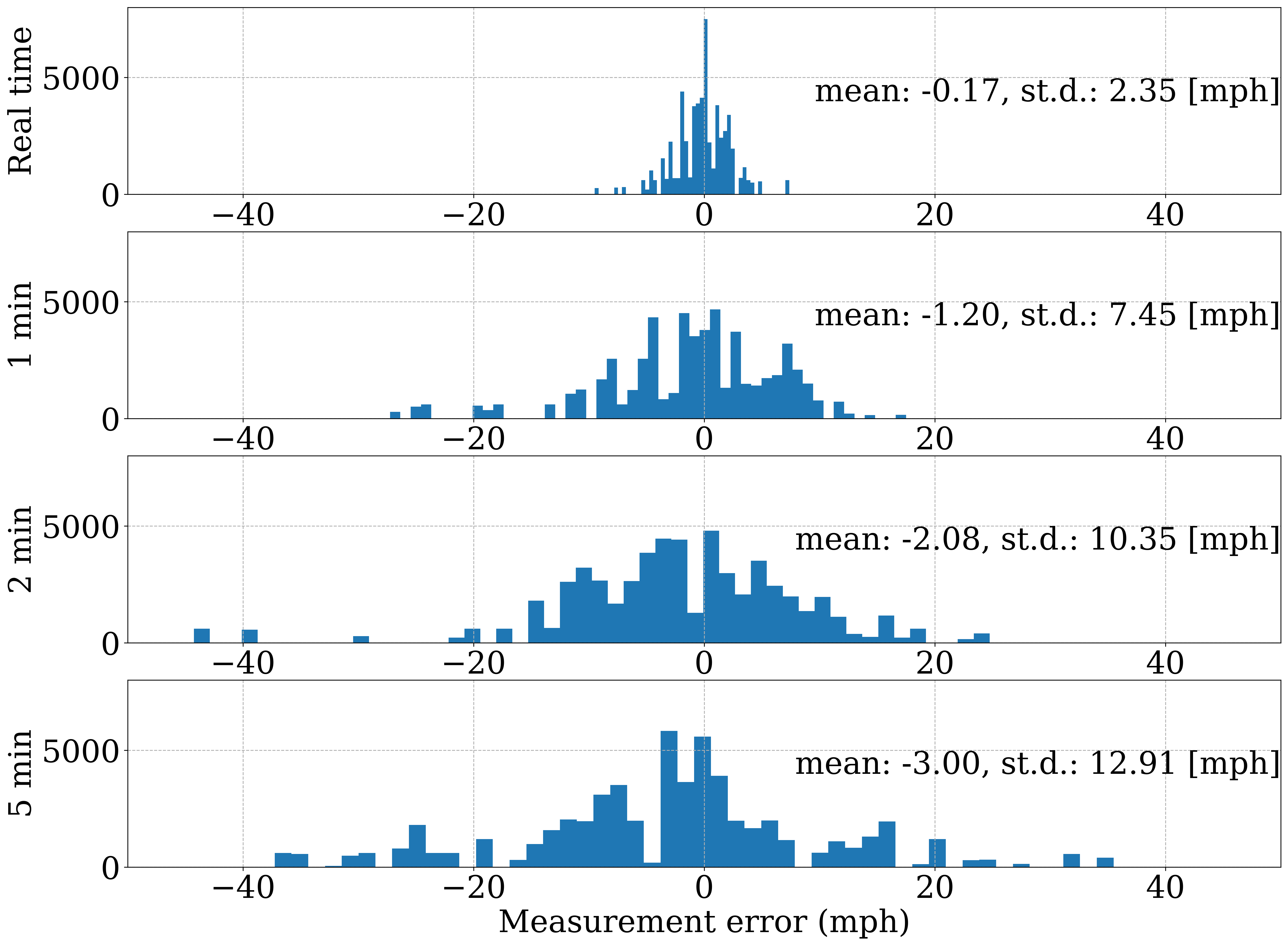}
    \caption{\textbf{Distribution of Latency-Induced Errors}: From Real time, to 1,2, and 5 minute latency. Notice a widening distribution of error in the measurement of local traffic speed. }
    \label{fig:latency}
\end{figure}

We subsequently compare the ideal speed measurement to the speed measurements that would have either been available in real-time, or with a certain amount of latency. Real-time speed estimates are calculated as the average only in space between the two RDS measurements most recently available at each trajectory point (but does not consider the 2 points ahead in time, as the ideal speed measurement does).  Additionally, we account for possible latency by shifting the trajectory only in time by a certain added latency, and then performing this calculation again. The errors between the real-time, 1 minute latency, 2 minute latency, and 5 minute latency speed estimates and that of the ideal speed measurements are shown in Figure~\ref{fig:speed_error}. In Figure~\ref{fig:latency} these errors are then shown as distributions. It is evident that real time RDS measurements have some error, and that latency in their measurement noticeably exacerbates the errors. The standard deviation of error increases $3.2$ times from 2.35 mph away from `ideal' to 7.45 mph with just one minute of delay. The standard deviations increase further, eclipsing over 10 mph of error, with standard deviation of 10.35 mph at two minutes of delay, and 12.91 mph at five minutes of delay. Commercial entities sell access to average traffic speeds with multi-minute delays, and the RDS sensors are limited to reporting over 30 second intervals. Considering the observation from Section~\ref{sec:motive} that in congested regions it is common to see 30 mph+ changes within 30 seconds, these fixed delay costs could be problematic. To avoid these issues, our control design opts to take estimates of the traffic speed 
from the nearby vehicles as measured by the on-board stock radar sensor.


\subsection{Controller Performance}
Earlier in the results we cover analyses informing design choices made to create a controller which is aware of the not just the variable speed limit, but also the local traffic speed. 
This subsection of the results showcases that controller's performance in deployment on the interstate during heavy morning congestion.

\begin{figure}
    \centering
    \includegraphics[width=\linewidth]{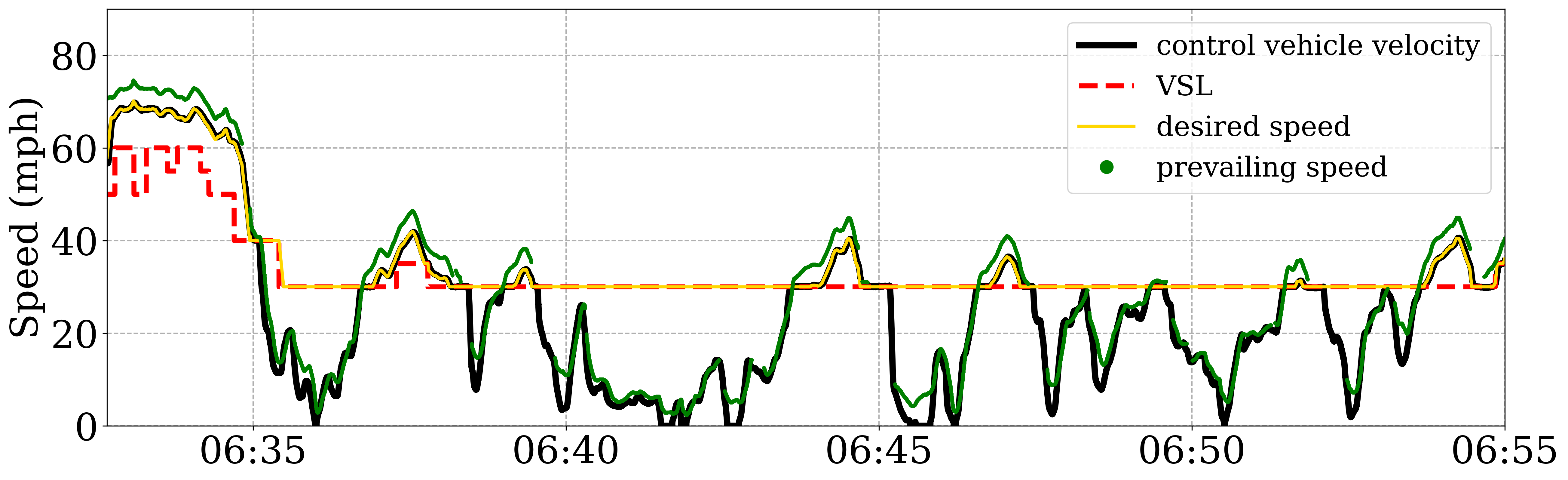}
    \caption{\textbf{A Single Complete Control Vehicle Trajectory}: An overview of a pass going through heavy morning congestion from the perspective of the control vehicle. $59.4\%$ of the time is spent in CBF-Mode. Entering congestion and at the peak of recurring waves, the vehicle is in VSL-Mode ($24.0\%$ of time) and Middleway-Mode ($16.6\%$ of time).
    }
    \label{fig:ControlVehTrajectory}
\end{figure}

 Figure~\ref{fig:ControlVehTrajectory} provides an overview of a driving trajectory with the novel controller. This plot emphasizes control decisions from the single vehicle perspective. In red, we show the posted VSL $v_{g_{r}}$. In green, we show the prevailing traffic $v_{pr}$. In gold, we show the desired speed $v_{des}$ from the controller. Speeds are initially near 60 mph ($v_{g_{r}}$) and 75 mph ($v_{pr}$); at $\sim$06:33 we see the VSL drop gradually to 30 mph. Note that $v_{g_{r}}$ is substantially slower than $v_{pr}$ until the speed 
 of traffic slows down to a stop at approximately 06:36. The control vehicle then speeds up and slows down over 20 more minutes in traffic waves.  There are three states predominantly driving the vehicle's velocity (black). (1) any time the vehicle velocity is below the $v_{g_{r}}$, the CBF-Mode safety control is active; (2) matching $v_{g_{r}}$ exactly (VSL-Mode); and (3) the speed 
 of traffic $v_{pr}$ is far enough above the $v_{g_{r}}$ that the vehicle deviates from the posted speed limit. Over the time periods where control was active in the deployment, $59.4\%$ of the time is spent in CBF-Mode, $24.0\%$ of the time is spent in VSL-Mode and $16.6\%$ of the time is spent in Middleway-Mode. In the entry to congestion before 06:35, the control vehicle is primarily keeping up with traffic, then pauses for a moment at $v_{g_{r}}$, before entering CBF-Mode. Throughout the rest of the congestion region on this westbound I-24 pass, the predominant driving automation is within the CBF-Mode; however, at the peak of the recurring traffic waves there are repeating opportunities for the novel controller to either travel in VSL-Mode, or speed up more to keep up with traffic speed 
 in Middleway-Mode. 

 \begin{figure}
    \centering
    \includegraphics[width=\linewidth]{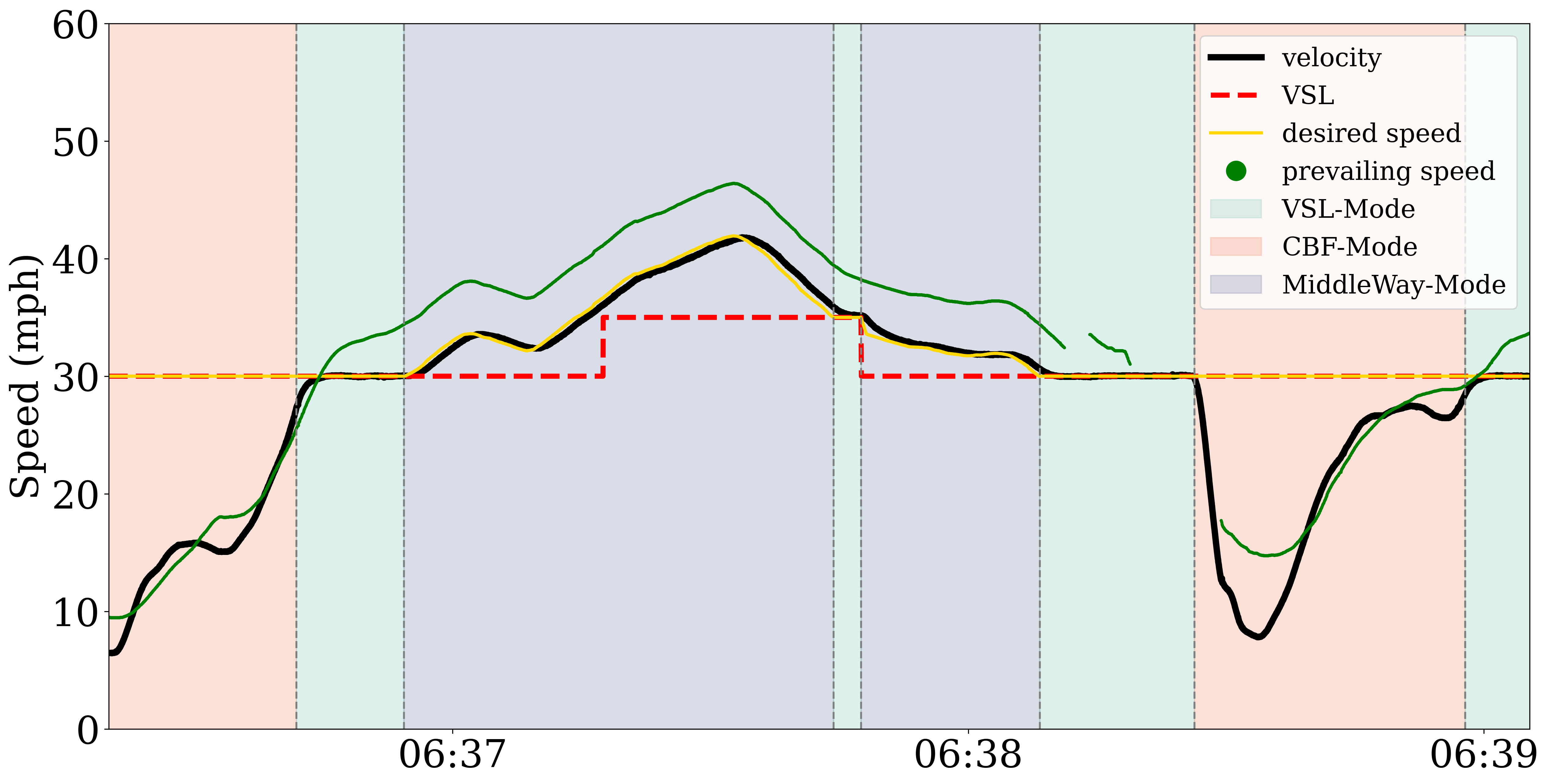}
    \caption{\textbf{Control in a Traffic Wave}: In a single traffic wave, we can understand the evolution of the state of the experimental control system in these recurring scenarios. Repeatedly, there traffic speeds up enough to allow the choice to follow the speed limit (VSL-Mode), then possibly speeds up faster than \veloffset above the speed limit inducing MiddleWay-Mode, and slows again to find VSL-Mode again, then well below the speed limit inducing CBF-mode.}
    \label{fig:modes}
\end{figure}
 
 Figure~\ref{fig:modes} gives a closer look at the behavior of the controller in the recurring traffic wave scenario. There are several transitions between control modes, which are highlighted in different colors. Time regions in red have CBF-Mode active, time regions in green are when adhering directly to the posted VSL (VSL-Mode), and time regions in purple are speeding above VSL due to faster traffic speed (Middleway-Mode).
 This plot begins at the end of slowest part of a traffic wave, where the control vehicle speeds up and has the opportunity to match the VSL. The CBF intervenes with limits on acceleration to prevent forward collisions, so once the preceding local traffic speeds up the control vehicle reaches $v_{des}$ at the posted VSL before 06:37. Approaching the velocity peak of the traffic wave, the traffic speed 
 is faster than the runtime offset parameter \veloffset ~(in this case $v_{pr}-2 \frac{m}{s}$), and the MiddleWay-Mode activates. Without this feature in place, a VSL-Mode following vehicle would have to weather a minute or so at each wave peak going 10-15 mph slower than traffic speed;
 this is an uncomfortable condition as a passenger. 

 In the wave shown in Figure~\ref{fig:modes} $v_{g_{r}}$ increases and the control vehicle catches the VSL-mode again at just before 06:38. Within a couple seconds the posted VSL $v_{g_{r}}$ drops down again, but instead of dropping down immediately the control vehicle stays within the offset below traffic speed 
 and eases over the next 30 seconds or so to the posted VSL setting as the traffic slows. Around 06:38:30 the CBF-Mode is activated again as the control vehicle reaches the entrance to the bottom of the traffic wave, and soon proceeds to the beginning of the cycle again.



\begin{figure}
    \centering
    \includegraphics[width=0.9\linewidth]{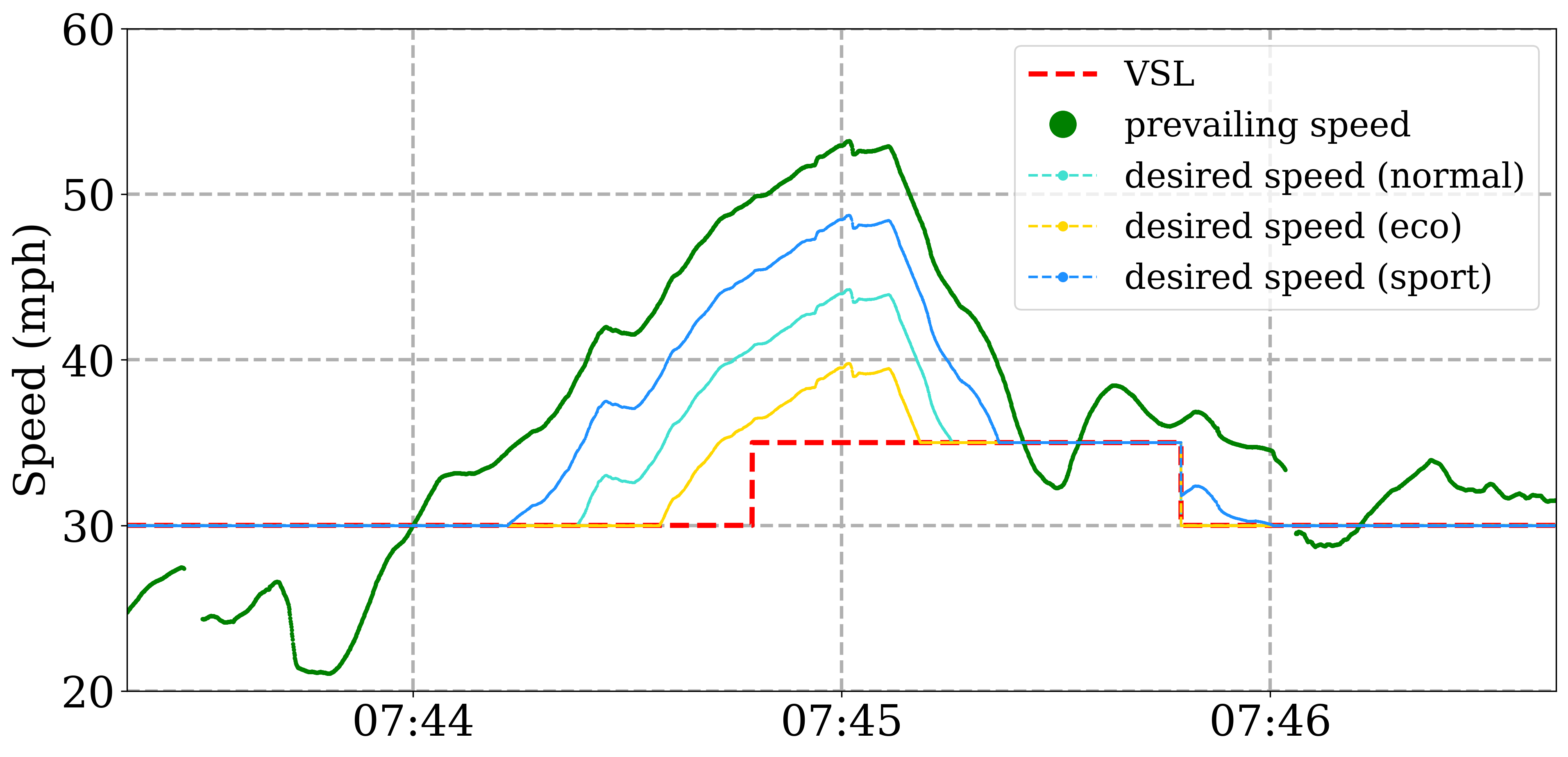}
    \caption{\textbf{Runtime Parameter \veloffset:} The offset from $v_{pr}$ can be set to 2/4/6 $\frac{m}{s}$ by the vehicle operator as they travel through congestion. This is achieved by listening to the vehicle drive mode (Sport/Normal/Eco, are 2/4/6 $\frac{m}{s}$ respectively). A smaller offset allows the novel controller to keep velocity closer to the local traffic speed,
    whereas a larger offset allows for more time travelling at the posted VSL.}
    \label{fig:offsets}
\end{figure}

In the recurring traffic wave scenario, the offset parameter \veloffset can be considered as the condition of how much faster than the posted speed limit does the vehicle need to observe the local traffic speed 
before deviating from the posted speed limit and speeding up. Figure~\ref{fig:offsets} shows an open loop projection of how different offsets would effect the desired velocity, $v_{des}$, with the same observed traffic speed,
$v_{pr}$, and posted VSL, $v_{g_{r}}$. At 07:44, when the plot for Figure~\ref{fig:offsets} begins, the prevailing speed is at $v_{g_{r}}$ so $v_{des}$ is at $v_{g_{r}} = 30$ mph for all offset settings. When $v_{pr} \le v_{g_{r}}$, $v_{des} = v_{g_{r}}$ by system definition. However, within a minute $v_{pr}$ has increased by over 20 mph ($\sim8\frac{m}{s}$). Now that $v_{pr} > v_{g_{r}}$ the offset parameter has a significant effect on where between $v_{pr}$ and $v_{g_{r}}$ the velocity of the vehicle will go. A smaller offset (i.e. `Sport Mode') leads to a quicker switch to tracking $v_{pr}$ instead of $v_{g_{r}}$. In the deployments made in this work, operators of the control vehicles chose primarily a `Sport Mode' $2\frac{m}{s}$ setting; when the `Default' $4\frac{m}{s}$ setting was in use, it was deemed too slow. This could vary in different traffic conditions and the difference between a larger population of operators; more investigation is needed to characterize \veloffset.

\section{Conclusions}
This paper presented a new cooperative automated vehicle controller that adopts variable speed limits set by smart infrastructure, and adapts to the speed of traffic when prevailing speeds warrant doing so for safety reasons, and relies on a control barrier function when following a vehicle ahead. The result was a new real-time algorithm which was enabled by high-latency communication to infrastructure, low-latency on-board sensors, and real-time algorithms on board the car that are informed by the cyber-physical properties of the ego car and the vehicles around it.

The maximum vehicle speed never exceeds the value specified by the driver using the heads-up-display, mitigating this source for mode confusion.
The updates to the variable speed limits can be made through mobile phone connectivity at high latency, without compromising the safety or efficacy of the solution. The implementation changes to deploy at scale are minimal, and do not require sensors or connectivity beyond what is present on most vehicles sold today. 

Field experiments validated the work on the open road during times of congestion when the VSL was active. The results show that the on-board ego car sensors were able to accurately estimate the speed of the flow of traffic (not just the speed of the ego car), as validated by roadside sensors. The field experiments demonstrate that each of the three modes of the presented controller are active during the drive in substantial portions, validating that the speed adaptation novelty has merit. 

Additional validation in the field experiment showed that speed 
of traffic far exceeds the posted speed limits as the ego car approached stopped traffic. This mode, in particular, validates the middle way approach: driving slower to increase effective compliance of the VSL, but in a way that follows accepted safety guidelines. Further,  traffic speed estimates from roadside sensors were shown to be unsuitable for real-time safety feedback, with latencies in which they are currently available. 

Future work will explore large-scale simulations, high-resolution measurement of the influence on neighboring cars, and additional field deployments at scale.
The large-scale simulation will explore how design choices in our prototype system would work at higher penetration rates. Further field deployments at scale can 
measure the influence on other vehicles in the flow, 
allowing us to infer an effective compliance rate based on our own measurements, to build advanced models for broader application in other system designs.



\bibliographystyle{IEEEtran}
\bibliography{references}

\begin{thebibliography}{10}
\providecommand{\url}[1]{#1}
\csname url@samestyle\endcsname
\providecommand{\newblock}{\relax}
\providecommand{\bibinfo}[2]{#2}
\providecommand{\BIBentrySTDinterwordspacing}{\spaceskip=0pt\relax}
\providecommand{\BIBentryALTinterwordstretchfactor}{4}
\providecommand{\BIBentryALTinterwordspacing}{\spaceskip=\fontdimen2\font plus
\BIBentryALTinterwordstretchfactor\fontdimen3\font minus \fontdimen4\font\relax}
\providecommand{\BIBforeignlanguage}[2]{{%
\expandafter\ifx\csname l@#1\endcsname\relax
\typeout{** WARNING: IEEEtran.bst: No hyphenation pattern has been}%
\typeout{** loaded for the language `#1'. Using the pattern for}%
\typeout{** the default language instead.}%
\else
\language=\csname l@#1\endcsname
\fi
#2}}
\providecommand{\BIBdecl}{\relax}
\BIBdecl

\bibitem{lu2014review}
X.-Y. Lu and S.~E. Shladover, ``Review of variable speed limits and advisories: Theory, algorithms, and practice,'' \emph{Transportation Research Record}, vol. 2423, no.~1, pp. 15--23, 2014.

\bibitem{sergeVSL}
\BIBentryALTinterwordspacing
S.~P. Hoogendoorn, W.~Daamen, R.~G. Hoogendoorn, and J.~W. Goemans, ``Assessment of dynamic speed limits on freeway a20 near rotterdam, netherlands,'' \emph{Transportation Research Record}, vol. 2380, no.~1, pp. 61--71, 2013. [Online]. Available: \url{https://doi.org/10.3141/2380-07}
\BIBentrySTDinterwordspacing

\bibitem{papageorgiou2008effects}
M.~Papageorgiou, E.~Kosmatopoulos, and I.~Papamichail, ``Effects of variable speed limits on motorway traffic flow,'' \emph{Transportation Research Record}, vol. 2047, no.~1, pp. 37--48, 2008.

\bibitem{ma2016freeway}
J.~Ma, X.~Li, S.~Shladover, H.~A. Rakha, X.-Y. Lu, R.~Jagannathan, and D.~J. Dailey, ``Freeway speed harmonization,'' \emph{IEEE Transactions on Intelligent Vehicles}, vol.~1, no.~1, pp. 78--89, 2016.

\bibitem{stern2018dissipation}
R.~E. Stern, S.~Cui, M.~L. Delle~Monache, R.~Bhadani, M.~Bunting, M.~Churchill, N.~Hamilton, H.~Pohlmann, F.~Wu, B.~Piccoli \emph{et~al.}, ``Dissipation of stop-and-go waves via control of autonomous vehicles: Field experiments,'' \emph{Transportation Research Part C: Emerging Technologies}, vol.~89, pp. 205--221, 2018.

\bibitem{delle2022new}
M.~Delle~Monache, C.~Pasquale, M.~Barreau, and R.~Stern, ``New frontiers of freeway traffic control and estimation,'' in \emph{2022 IEEE 61st Conference on Decision and Control (CDC)}.\hskip 1em plus 0.5em minus 0.4em\relax IEEE, 2022, pp. 6910--6925.

\bibitem{grumert2015analysis}
E.~Grumert, X.~Ma, and A.~Tapani, ``Analysis of a cooperative variable speed limit system using microscopic traffic simulation,'' \emph{Transportation research part C: emerging technologies}, vol.~52, pp. 173--186, 2015.

\bibitem{khondaker2015variable}
B.~Khondaker and L.~Kattan, ``Variable speed limit: A microscopic analysis in a connected vehicle environment,'' \emph{Transportation Research Part C: Emerging Technologies}, vol.~58, pp. 146--159, 2015.

\bibitem{zhang2023marvel}
Y.~Zhang, M.~Quinones-Grueiro, Z.~Zhang, Y.~Wang, W.~Barbour, G.~Biswas, and D.~Work, ``Marvel: Multi-agent reinforcement-learning for large-scale variable speed limits,'' \emph{arXiv preprint arXiv:2310.12359}, 2023.

\bibitem{de2018safety}
E.~De~Pauw, S.~Daniels, L.~Franckx, and I.~Mayeres, ``Safety effects of dynamic speed limits on motorways,'' \emph{Accident Analysis \& Prevention}, vol. 114, pp. 83--89, 2018.

\bibitem{pu2020full}
Z.~Pu, Z.~Li, Y.~Jiang, and Y.~Wang, ``Full bayesian before-after analysis of safety effects of variable speed limit system,'' \emph{IEEE transactions on intelligent transportation systems}, vol.~22, no.~2, pp. 964--976, 2020.

\bibitem{hadiuzzaman2015modeling}
M.~Hadiuzzaman, J.~Fang, M.~A. Karim, Y.~Luo, and T.~Z. Qiu, ``Modeling driver compliance to vsl and quantifying impacts of compliance levels and control strategy on mobility and safety,'' \emph{Journal of transportation engineering}, vol. 141, no.~12, p. 04015028, 2015.

\bibitem{hellinga2011impact}
B.~Hellinga and M.~Mandelzys, ``Impact of driver compliance on the safety and operational impacts of freeway variable speed limit systems,'' \emph{Journal of transportation engineering}, vol. 137, no.~4, pp. 260--268, 2011.

\bibitem{habtemichael2013safety}
F.~G. Habtemichael and L.~de~Picado~Santos, ``Safety and operational benefits of variable speed limits under different traffic conditions and driver compliance levels,'' \emph{Transportation research record}, vol. 2386, no.~1, pp. 7--15, 2013.

\bibitem{zhang2022quantifying}
Y.~Zhang, M.~Quinones-Grueiro, W.~Barbour, C.~Weston, G.~Biswas, and D.~Work, ``Quantifying the impact of driver compliance on the effectiveness of variable speed limits and lane control systems,'' in \emph{2022 IEEE 25th International Conference on Intelligent Transportation Systems (ITSC)}, 2022, pp. 3638--3644.

\bibitem{nissan2011evaluation}
A.~Nissan and H.~N. Koutsopoulosb, ``Evaluation of the impact of advisory variable speed limits on motorway capacity and level of service,'' \emph{Procedia-Social and Behavioral Sciences}, vol.~16, pp. 100--109, 2011.

\bibitem{talebpour2016influence}
A.~Talebpour and H.~S. Mahmassani, ``Influence of connected and autonomous vehicles on traffic flow stability and throughput,'' \emph{Transportation research part C: emerging technologies}, vol.~71, pp. 143--163, 2016.

\bibitem{lochrane2014using}
T.~W. Lochrane, H.~Al-Deek, D.~J. Dailey, and J.~Bared, ``Using a living laboratory to support transportation research for a freeway work zone,'' \emph{Journal of Transportation Engineering}, vol. 140, no.~7, p. 04014024, 2014.

\bibitem{nice2023enabling}
M.~Nice, M.~Bunting, A.~Richardon, G.~Zachar, J.~W. Lee, A.~Bayen, M.~L. Delle~Monache, B.~Seibold, B.~Piccoli, J.~Sprinkle, and D.~Work, ``Enabling mixed autonomy traffic control,'' \emph{arXiv preprint arXiv:2310.18776}, 2023.

\bibitem{xiao2021rule}
W.~Xiao, N.~Mehdipour, A.~Collin, A.~Y. Bin-Nun, E.~Frazzoli, R.~D. Tebbens, and C.~Belta, ``Rule-based optimal control for autonomous driving,'' in \emph{Proceedings of the ACM/IEEE 12th International Conference on Cyber-Physical Systems}, 2021, pp. 143--154.

\bibitem{schwarting2019social}
W.~Schwarting, A.~Pierson, J.~Alonso-Mora, S.~Karaman, and D.~Rus, ``Social behavior for autonomous vehicles,'' \emph{Proceedings of the National Academy of Sciences}, vol. 116, no.~50, pp. 24\,972--24\,978, 2019.

\bibitem{nice2021can}
M.~Nice, S.~Elmadani, R.~Bhadani, M.~Bunting, J.~Sprinkle, and D.~Work, ``Can coach: vehicular control through human cyber-physical systems,'' in \emph{Proceedings of the ACM/IEEE 12th International Conference on Cyber-Physical Systems}, 2021, pp. 132--142.

\bibitem{bian2019reducing}
Y.~Bian, Y.~Zheng, W.~Ren, S.~E. Li, J.~Wang, and K.~Li, ``Reducing time headway for platooning of connected vehicles via v2v communication,'' \emph{Transportation Research Part C: Emerging Technologies}, vol. 102, pp. 87--105, 2019.

\bibitem{aoki2018dynamic}
S.~Aoki and R.~Rajkumar, ``Dynamic intersections and self-driving vehicles,'' in \emph{2018 ACM/IEEE 9th International Conference on Cyber-Physical Systems (ICCPS)}.\hskip 1em plus 0.5em minus 0.4em\relax IEEE, 2018, pp. 320--330.

\bibitem{li2017integrated}
Y.~Li, C.~Xu, L.~Xing, and W.~Wang, ``Integrated cooperative adaptive cruise and variable speed limit controls for reducing rear-end collision risks near freeway bottlenecks based on micro-simulations,'' \emph{IEEE transactions on intelligent transportation systems}, vol.~18, no.~11, pp. 3157--3167, 2017.

\bibitem{nice2023sailing}
M.~Nice, M.~Bunting, G.~Gunter, W.~Barbour, J.~Sprinkle, and D.~Work, ``Sailing cavs: Speed-adaptive infrastructure-linked connected and automated vehicles,'' \emph{arXiv preprint arXiv:2310.06931}, 2023.

\bibitem{bunting2021libpanda}
M.~Bunting, R.~Bhadani, and J.~Sprinkle, ``Libpanda: A high performance library for vehicle data collection,'' in \emph{Proceedings of the Workshop on Data-Driven and Intelligent Cyber-Physical Systems}, 2021, pp. 32--40.

\bibitem{gunter2022experimental}
G.~Gunter, M.~Nice, M.~Bunting, J.~Sprinkle, and D.~B. Work, ``Experimental testing of a control barrier function on an automated vehicle in live multi-lane traffic,'' in \emph{2022 2nd Workshop on Data-Driven and Intelligent Cyber-Physical Systems for Smart Cities Workshop (DI-CPS)}.\hskip 1em plus 0.5em minus 0.4em\relax IEEE, 2022, pp. 31--35.

\bibitem{gunter2022CBFs_for_cutins}
G.~Gunter and D.~Work, ``Safe driving with control barrier functions in mixed autonomy traffic when cut-ins occur,'' in \emph{2022 European Control Conference (ECC)}, 2022, pp. 411--416.

\bibitem{ames2016control}
A.~D. Ames, X.~Xu, J.~W. Grizzle, and P.~Tabuada, ``Control barrier function based quadratic programs for safety critical systems,'' \emph{IEEE Transactions on Automatic Control}, vol.~62, no.~8, pp. 3861--3876, 2016.

\bibitem{ames2019control}
A.~D. Ames, S.~Coogan, M.~Egerstedt, G.~Notomista, K.~Sreenath, and P.~Tabuada, ``Control barrier functions: Theory and applications,'' in \emph{2019 18th European control conference (ECC)}.\hskip 1em plus 0.5em minus 0.4em\relax IEEE, 2019, pp. 3420--3431.

\bibitem{quigley2009ros}
M.~Quigley, K.~Conley, B.~Gerkey, J.~Faust, T.~Foote, J.~Leibs, R.~Wheeler, and A.~Ng, ``{ROS: an open-source Robot Operating System},'' in \emph{ICRA workshop on open source software}, ser. 1, no. 3.2.\hskip 1em plus 0.5em minus 0.4em\relax Kobe, Japan: ICRA, 2009, p.~5.

\bibitem{nice2023middleware}
M.~W. Nice, M.~Bunting, J.~Sprinkle, and D.~Work, ``Middleware for a heterogeneous cav fleet,'' in \emph{2023 5th Workshop on Design Automation for CPS and IoT (DESTION)}, 2023.

\bibitem{elmadani2021can}
S.~Elmadani, M.~Nice, M.~Bunting, J.~Sprinkle, and R.~Bhadani, ``From can to ros: A monitoring and data recording bridge,'' in \emph{Proceedings of the Workshop on Data-Driven and Intelligent Cyber-Physical Systems}, 2021, pp. 17--21.

\bibitem{tdot2023i24}
\BIBentryALTinterwordspacing
T.~D. of~Transportation, ``Interstate 24 smart corridor.'' [Online]. Available: \url{https://www.tn.gov/tdot/projects/region-3/i-24-smart-corridor.html}
\BIBentrySTDinterwordspacing

\end{thebibliography}

\clearpage








\end{document}